\documentclass[11pt]{article}

\usepackage[utf8]{inputenc}
\usepackage[T1]{fontenc}
\usepackage{mathpazo} 
\usepackage{microtype} 
\usepackage[left=1in, right=1in, top=0.8in, bottom=1in]{geometry}
\usepackage{listings}
\usepackage{graphicx}
\usepackage{xcolor}
\usepackage{booktabs}
\usepackage{siunitx}
\usepackage{pgfplots}
\usepackage{hyperref}
\usepackage{caption}
\usepackage{subcaption}
\usepackage{float}
\usepackage{titlesec}
\usepackage{amsmath}
\usepackage{wrapfig}

\pgfplotsset{compat=1.18}

\definecolor{asterdark}{HTML}{7879b6} 
\definecolor{diffstart}{named}{gray}
\definecolor{diffincl}{HTML}{006400} 
\definecolor{diffrem}{HTML}{8B0000}  

\usepackage[most]{tcolorbox}
\tcbuselibrary{listings}

\lstdefinelanguage{Diff}{
    basicstyle=\ttfamily\scriptsize,
    morecomment=[f][\color{diffstart}]{@@},
    morecomment=[f][\color{diffincl}]{+\ },
    morecomment=[f][\color{diffincl}]{+},
    morecomment=[f][\color{diffrem}]{-\ },
    morecomment=[f][\color{diffrem}]{-},
    keepspaces=true,
    breaklines=true,
    columns=fullflexible,
}

\newtcblisting{diffbox}[1]{
    enhanced,
    breakable,               
    colback=white,
    colframe=asterdark,      
    colbacktitle=asterdark,  
    fonttitle=\bfseries,
    coltitle=white,
    title={#1},
    boxrule=1pt,             
    arc=1mm,                 
    listing only,
    listing options={
        language=Diff,
        basicstyle=\ttfamily\scriptsize,
        breaklines=true,
    },
}

\newcommand{\name}{TTT-Discover}

\definecolor{pastelblue}{HTML}{E3F2FD}
\definecolor{pastelgreen}{HTML}{E8F5E9}
\definecolor{pastellavender}{HTML}{EDE7F6}
\definecolor{pastelpeach}{HTML}{FFF3E0}
\definecolor{pastelrose}{HTML}{FCE4EC}
\definecolor{pastelmint}{HTML}{E0F2F1}
\usepackage{fontawesome5} 
\newtcolorbox{reviewbox}[2][]{%
  enhanced,
  breakable,
  colback=#2,
  colframe=#2!60!black!20,
  boxrule=0pt,
  arc=4mm,
  outer arc=4mm,
  left=5mm, right=5mm, top=4mm, bottom=4mm,
  shadow={1.5mm}{-1.5mm}{0mm}{black!8},
  attach boxed title to top left={yshift=-3mm, xshift=5mm},
  boxed title style={
    colback=white,
    colframe=white,
    arc=2mm,
    outer arc=2mm,
    boxrule=0pt,
    left=2mm, right=2mm, top=1mm, bottom=1mm,
    shadow={1mm}{-1mm}{0mm}{black!6},
  },
  fonttitle=\normalsize\bfseries,
  coltitle=black,
  title={Human Expert Review~---~#1},
}

\definecolor{artifactbg}{HTML}{ffffff}
\definecolor{artifactframe}{HTML}{d0d0d0}
\definecolor{artifacttitle}{HTML}{333333}

\definecolor{asterdark}{HTML}{7879b6}

\definecolor{artifactbg}{HTML}{ffffff}
\definecolor{artifactframe}{HTML}{d0d0d0}
\definecolor{artifacttitle}{HTML}{333333}

\definecolor{codegreen}{HTML}{228B22}
\definecolor{codeblue}{HTML}{0000CD}
\definecolor{codegray}{HTML}{808080}

\definecolor{promptbg}{HTML}{f9f6f2}
\definecolor{promptframe}{HTML}{d4c4b0}
\definecolor{promptaccent}{HTML}{5a4a3a}

\usepackage{upquote}
\usepackage{listingsutf8}
\lstdefinestyle{artifactstyle}{
    basicstyle=\ttfamily\scriptsize,
    backgroundcolor=\color{artifactbg},
    commentstyle=\color{codegray}\itshape,
    keywordstyle=\color{codeblue},
    stringstyle=\color{codegreen},
    numberstyle=\tiny\color{codegray},
    breakatwhitespace=false,
    breaklines=true,
    keepspaces=true,
    numbers=none,
    showspaces=false,
    showstringspaces=false,
    showtabs=false,
    tabsize=4,
    xleftmargin=12pt,
    framexleftmargin=12pt,
    aboveskip=0pt,
    belowskip=0pt,
    inputencoding=utf8,
    columns=fullflexible,
    upquote=true,
    literate=
        {→}{{$\rightarrow$}}1
        {←}{{$\leftarrow$}}1
        {↔}{{$\leftrightarrow$}}1
        {≤}{{$\leq$}}1
        {≥}{{$\geq$}}1
        {≠}{{$\neq$}}1
        {−}{{-}}1
        {—}{{--}}1
        {–}{{-}}1
        {"}{{\textquotedblleft}}1
        {"}{{\textquotedblright}}1
        {'}{{\textquoteleft}}1
        {'}{{\textquoteright}}1
        {…}{{...}}1
        {×}{{$\times$}}1
        {÷}{{$\div$}}1
        {±}{{$\pm$}}1
        {∞}{{$\infty$}}1
        {α}{{$\alpha$}}1
        {β}{{$\beta$}}1
        {γ}{{$\gamma$}}1
        {δ}{{$\delta$}}1
        {ε}{{$\epsilon$}}1
        {λ}{{$\lambda$}}1
        {π}{{$\pi$}}1
        {σ}{{$\sigma$}}1
        {∑}{{$\sum$}}1
        {∏}{{$\prod$}}1
        {√}{{$\sqrt{}$}}1
        {∈}{{$\in$}}1
        {∉}{{$\notin$}}1
        {⊂}{{$\subset$}}1
        {⊃}{{$\supset$}}1
        {∩}{{$\cap$}}1
        {∪}{{$\cup$}}1
        {∀}{{$\forall$}}1
        {∃}{{$\exists$}}1
        {¬}{{$\neg$}}1
        {∧}{{$\land$}}1
        {∨}{{$\lor$}}1,
}

\lstdefinestyle{promptstyle}{
    basicstyle=\small\scriptsize,
    backgroundcolor=\color{promptbg},
    breakatwhitespace=false,
    breaklines=true,
    breakindent=0pt,
    breakautoindent=false,
    keepspaces=true,
    numbers=none,
    showspaces=false,
    showstringspaces=false,
    showtabs=false,
    tabsize=4,
    aboveskip=0pt,
    belowskip=0pt,
    xleftmargin=0pt,
    framexleftmargin=0pt,
    extendedchars=true,
    inputencoding=utf8,
    upquote=true,
    literate=
        {→}{{$\rightarrow$}}1
        {←}{{$\leftarrow$}}1
        {↔}{{$\leftrightarrow$}}1
        {≤}{{$\leq$}}1
        {≥}{{$\geq$}}1
        {≠}{{$\neq$}}1
        {−}{{-}}1
        {—}{{--}}1
        {–}{{-}}1
        {"}{{\textquotedblleft}}1
        {"}{{\textquotedblright}}1
        {'}{{\textquoteleft}}1
        {'}{{\textquoteright}}1
        {…}{{...}}1
        {×}{{$\times$}}1
        {÷}{{$\div$}}1
        {±}{{$\pm$}}1
        {∞}{{$\infty$}}1
        {α}{{$\alpha$}}1
        {β}{{$\beta$}}1
        {γ}{{$\gamma$}}1
        {δ}{{$\delta$}}1
        {ε}{{$\epsilon$}}1
        {λ}{{$\lambda$}}1
        {π}{{$\pi$}}1
        {σ}{{$\sigma$}}1
        {∑}{{$\sum$}}1
        {∏}{{$\prod$}}1
        {√}{{$\sqrt{}$}}1
        {∈}{{$\in$}}1
        {∉}{{$\notin$}}1
        {⊂}{{$\subset$}}1
        {⊃}{{$\supset$}}1
        {∩}{{$\cap$}}1
        {∪}{{$\cup$}}1
        {∀}{{$\forall$}}1
        {∃}{{$\exists$}}1
        {¬}{{$\neg$}}1
        {∧}{{$\land$}}1
        {∨}{{$\lor$}}1,
}

\newtcblisting{artifact}[2][python]{%
    enhanced,
    breakable,
    colback=artifactbg,
    colframe=artifactframe,
    boxrule=0.5pt,
    arc=3mm,
    outer arc=3mm,
    left=0mm, right=3mm, top=2mm, bottom=2mm,
    shadow={1.5mm}{-1.5mm}{0mm}{black!10},
    attach boxed title to top left={yshift=-3mm, xshift=5mm},
    boxed title style={
        colback=white,
        colframe=artifactframe,
        arc=2mm,
        outer arc=2mm,
        boxrule=0.5pt,
        left=2mm, right=2mm, top=1mm, bottom=1mm,
    },
    fonttitle=\normalsize\bfseries,
    coltitle=artifacttitle,
    title={#2},
    listing only,
    listing options={
        style=artifactstyle,
        language=#1,
    },
}

\newtcblisting{prompt}[1]{%
    enhanced,
    breakable,
    colback=promptbg,
    colframe=promptframe,
    boxrule=0.5pt,
    arc=3mm,
    outer arc=3mm,
    left=5mm, right=5mm, top=4mm, bottom=4mm,
    shadow={1.5mm}{-1.5mm}{0mm}{black!8},
    attach boxed title to top left={yshift=-3mm, xshift=5mm},
    boxed title style={
        colback=white,
        colframe=promptframe,
        arc=2mm,
        outer arc=2mm,
        boxrule=0.5pt,
        left=2mm, right=2mm, top=1mm, bottom=1mm,
    },
    fonttitle=\normalsize,
    coltitle=promptaccent,
    title={#1},
    listing only,
    listing options={
        style=promptstyle,
    },
}
\renewcommand{\name}{Aster} 

\titlespacing*{\section}{0pt}{0.8em}{0.4em}
\titlespacing*{\subsection}{0pt}{0.6em}{0.3em}

\begin{document}

\noindent
\begin{minipage}[b]{0.1\textwidth}
    \includegraphics[width=2.5cm]{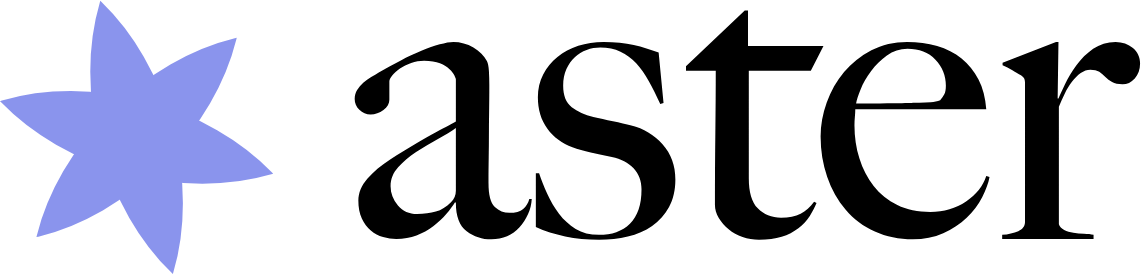} 
\end{minipage}
\hfill
\begin{minipage}[b]{0.5\textwidth}
    \raggedleft
    \textcolor{gray}{\large \today} 
\end{minipage}

\vspace{0.2em}
\hrule height 1.5pt 
\vspace{1.2em}

\begin{center}
    {\fontsize{20}{24}\selectfont \textbf{Aster: Autonomous Scientific Discovery \\ over 20x Faster Than Existing Methods}}
    
    \vspace{0.8em}
    {\large \textbf{Emmett Bicker}}
    
    \vspace{0.2em}
    {\footnotesize $^{}$Aster AI Labs Inc., San Francisco, CA, USA} \\
    {\footnotesize \texttt{emmett@asterlab.ai}}
    
    \vspace{1.2em}
    
    \begin{minipage}{0.9\textwidth}
        \centering
        \textbf{\normalsize Abstract}
        \vspace{0.4em} 
        
        \raggedright 
        \setlength{\parindent}{1em}  
        \small 
        We introduce \name{}, an AI agent for autonomous scientific discovery capable of operating over 20x faster than existing frameworks. Given a task, an initial program, and a script to evaluate the performance of the program, \name{} iteratively improves the program, often leading to new state-of-the-art performances. \name{}'s significant reduction in the number of iterations required for novel discovery expands the domain of tractable problems to include tasks with long evaluation durations, such as multi-hour machine learning training runs.
        \\
        We applied \name{} to problems in mathematics, GPU kernel engineering, biology, neuroscience, and language model training. More specifically: the Erdős’ minimum overlap problem, optimizing the TriMul kernel, a single-cell analysis denoising problem, training a neural activity prediction model to perform well on ZAPBench, and the NanoGPT Speedrun Competition. \name{} attains SOTA results in every task, except for ZAPBench, where it matches the performance of the best human solution with < 1/190th of the compute. 
        
        \name{} is accessible via a web interface and API at \href{https://www.asterlab.ai}{asterlab.ai}.
    \end{minipage}
\end{center}

\vspace{1em}
\begin{figure}[h!]
    \centering
    \small
    \setlength{\tabcolsep}{10pt}
    \renewcommand{\arraystretch}{1.2}
    \begin{tabular}{@{}lcccccc@{}}
    \toprule
    & \textbf{Math} 
    & \multicolumn{2}{c}{\textbf{Kernel Eng.}} 
    & \textbf{Biology}
    & \textbf{ML} \\
    & \scriptsize{Overlap ($\downarrow$)} 
    & \scriptsize{TriMul ($\downarrow$)} 
    & 
    & \scriptsize{Denoise ($\uparrow$)}
    & \scriptsize{NanoGPT ($\downarrow$)} \\
    \midrule
    Best Human
    & $0.380927${\scriptsize\cite{haugland2016minimum}}
    & \SI{1371}{\micro\second} 
    &
    & $0.641$
    & $96.8${\scriptsize\cite{nanogpt_speedrun}} \\
    Prev. Best AI
    & $0.380876$
    & \SI{1161}{\micro\second}
    &
    & $0.709$
    & N/A \\
    \midrule
    \textbf{\textcolor{asterdark}{\name{}}}
    & \textcolor{asterdark}{$\mathbf{0.380874}$} 
    & \textcolor{asterdark}{\textbf{\SI{1114}{\micro\second}}}
    &
    & \textcolor{asterdark}{$\mathbf{0.711}$}
    & \textcolor{asterdark}{$\mathbf{95.2}$} \\
    \bottomrule
    \end{tabular}
    \captionof{table}{Discoveries found by \name{} across Mathematics, Kernel Engineering, Biology, and ML. All baseline results cited from TTT-Discover\cite{yuksekgonul2026learningdiscovertesttime} unless cited otherwise.}
    \vspace{2.5em} 
    \begin{minipage}{0.5\textwidth}
        \centering
        \begin{tikzpicture}[scale=0.85]
        \begin{axis}[
            ybar,
            width=8cm,
            height=4.5cm,
            bar width=1.8cm,
            bar shift=0pt,
            symbolic x coords={Aster, OpenEvolve},
            xtick={Aster, OpenEvolve},
            nodes near coords={\pgfmathprintnumber[fixed, precision=0]{\pgfplotspointmeta}x},
            nodes near coords style={font=\bfseries\small, yshift=2pt},
            axis lines=left,
            ymajorticks=false,
            ylabel={},
            ymin=0, ymax=30,
            title={Iteration Speedup (Circle Packing)},
            title style={font=\large\bfseries},
            enlarge x limits=0.6,
            axis line style={-},
            tick style={draw=none}
        ]
            \addplot[fill=asterdark, draw=black] coordinates {(Aster, 23)};
            \addplot[fill=blue!10!white, draw=black] coordinates {(OpenEvolve, 1.0)};
        \end{axis}
        \end{tikzpicture}
    \end{minipage}
    \begin{minipage}{0.4\textwidth}
        \centering
        \small
        \renewcommand{\arraystretch}{1.4}
        \begin{tabular}{lc}
            \toprule
            \textbf{Method} & \textbf{Iterations} \\
            \midrule
            \textbf{\name{}} & \textbf{5} \\
            OpenEvolve & 115 \\
            \bottomrule
        \end{tabular}
        \vspace{1.5em}
    \end{minipage}
    \captionof{figure}{A depiction of \name{} converging over 20x faster than OpenEvolve}
\end{figure}

\vspace{1em}

\section*{1. Introduction}
Autonomous discovery systems are transforming research by using Large Language Models (LLMs) to iteratively improve code against user-defined objectives. These systems have already yielded significant breakthroughs, such as discovering a new algorithm for multiplying 4x4 complex matrices for the first time in decades\cite{alphaevolve}, engineering high-performance kernels for machine learning systems\cite{cheng2025barbarians}, and creating new records for mathematical constructions\cite{romera2024mathematical, alphaevolve}.

\begin{figure}[t!]
    \centering
    \includegraphics[width=0.7\linewidth]{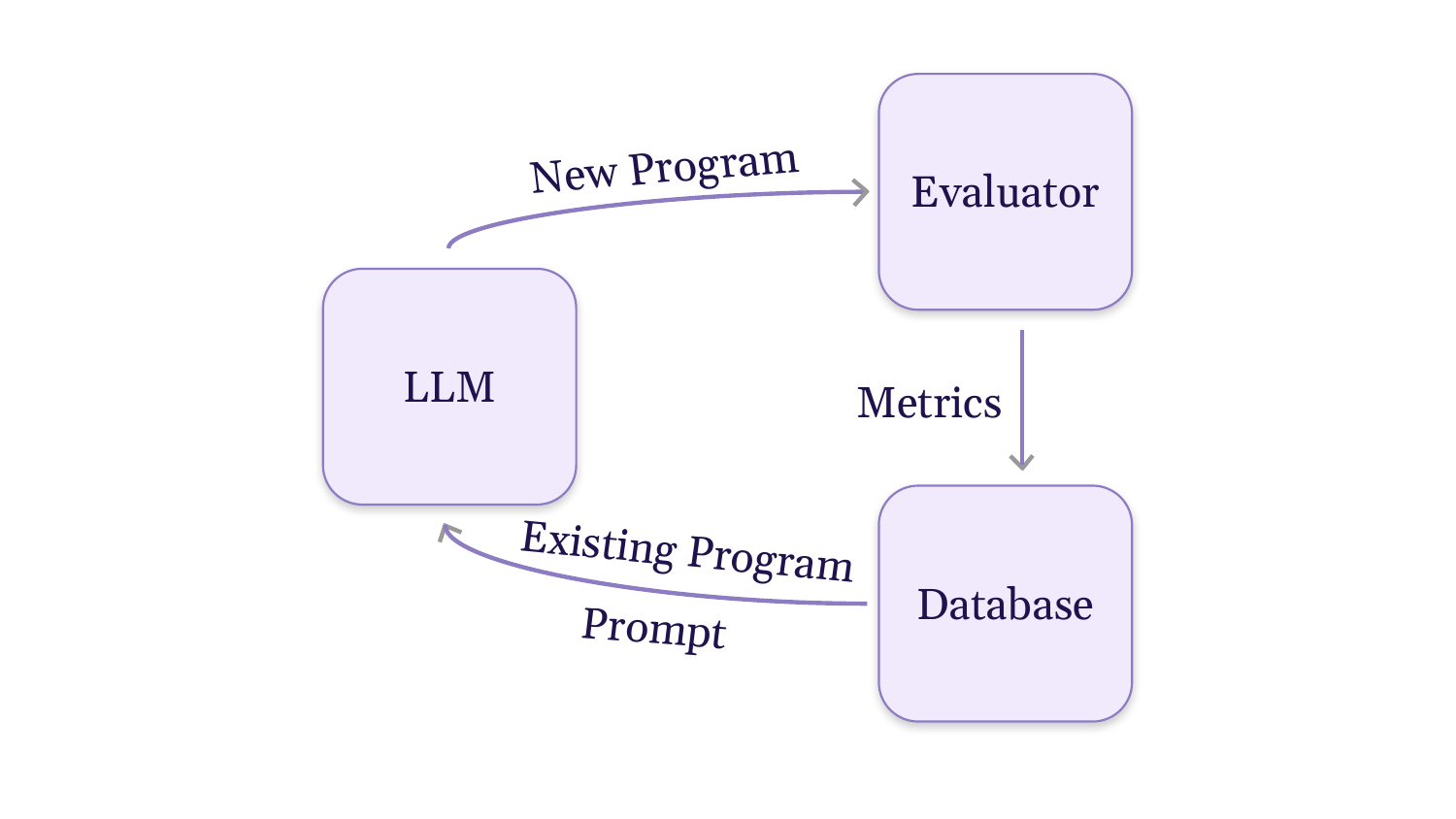}
    \caption{System Overview: \name{} iteratively refines programs, evaluates their performance, stores them in a database, and repeats.}
    \label{fig:system_overview}
\end{figure}

However, current state-of-the-art frameworks often require hundreds to thousands of iterations to achieve a discovery\cite{shinkaevolve, openevolve, alphaevolve}. This inefficiency restricts their application to problems with short evaluation times. For tasks requiring long-duration evaluations—such as training large machine learning models—the time to run thousands of iterations becomes prohibitive.

\name{} addresses this bottleneck. It operates as an autonomous agent similar to existing frameworks\cite{openevolve, shinkaevolve}, taking an initial program, an evaluator, and a prompt to iteratively refine the solution. Critically, \name{} achieves an \textbf{over 20x speedup} compared to leading open-source discovery system OpenEvolve, enabling it to tackle computationally intensive discovery tasks previously out of reach.

\section*{2. Speedup Analysis}
We benchmarked \name{}'s efficiency on the circle packing problem (packing 26 circles), a standard task in the field of autonomous discovery\cite{openevolve, shinkaevolve, wang2025thetaevolve}. We benchmark \name{} against OpenEvolve, as it's a prominent open-source framework for autonomous discovery.

To ensure a fair comparison, we configured \name{} to use the same underlying model distribution as OpenEvolve (80\% Gemini 2.0 Flash, 20\% Claude 3.7 Sonnet). OpenEvolve attempted the problem 460 times, but since it generated programs 4x in parallel, we count it as performing 115 iterations. While OpenEvolve required 115 iterations to reach a score of 2.634, \name{} surpassed this threshold with a score of \textbf{2.6353} in just \textbf{5 iterations}. Additionally, the current known state-of-the-art for this problem is \textbf{2.635983}; \name{}, utilizing its default model configuration, reaches this SOTA result in only \textbf{6 iterations}.

\begin{figure}[h!]
    \centering
    \begin{subfigure}{0.40\textwidth}
        \centering
        \includegraphics[width=\linewidth]{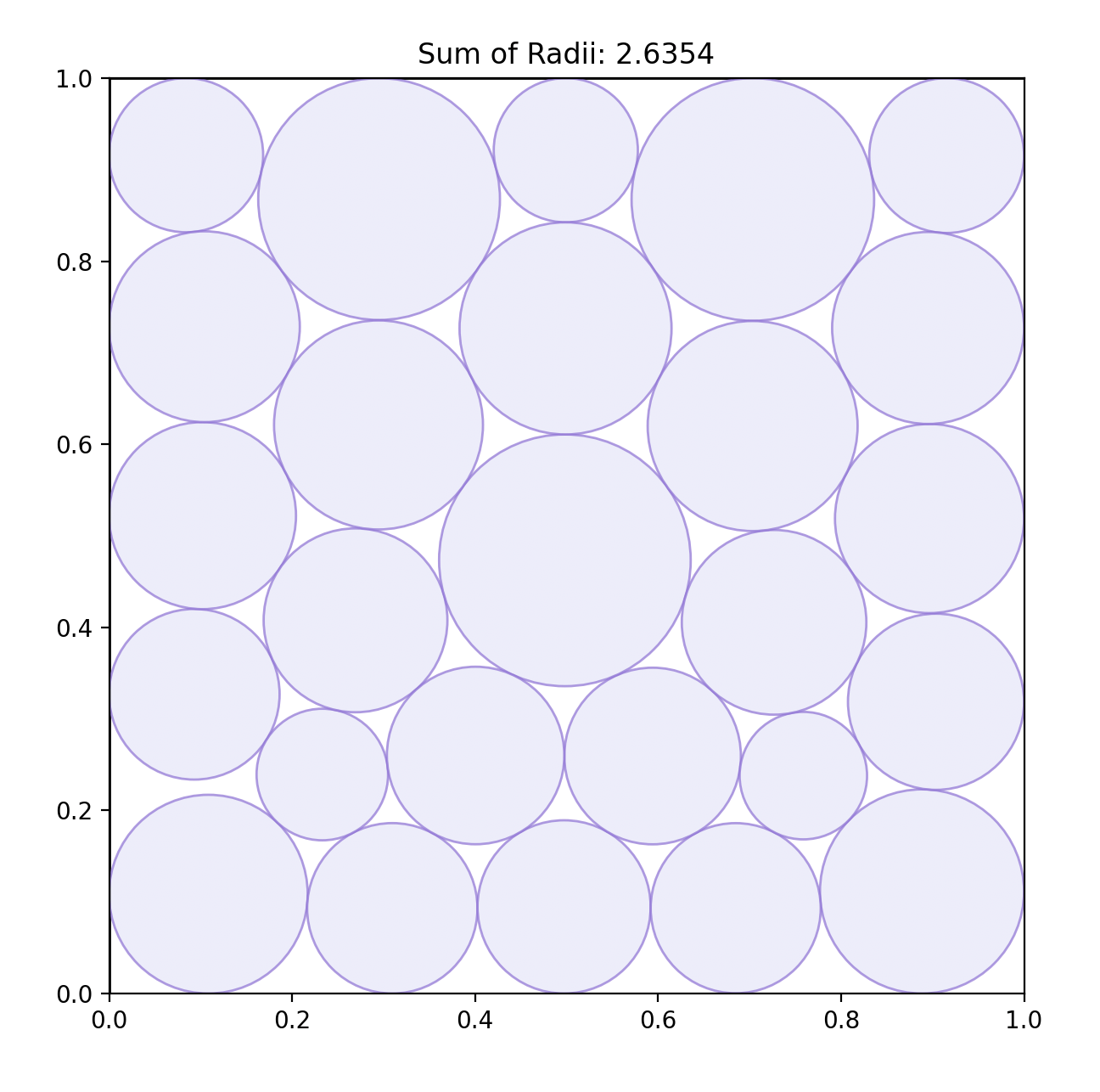}
        \caption*{\name{}: Iteration 5}
    \end{subfigure}
    \hfill
    \begin{subfigure}{0.40\textwidth}
        \centering
        \includegraphics[width=\linewidth]{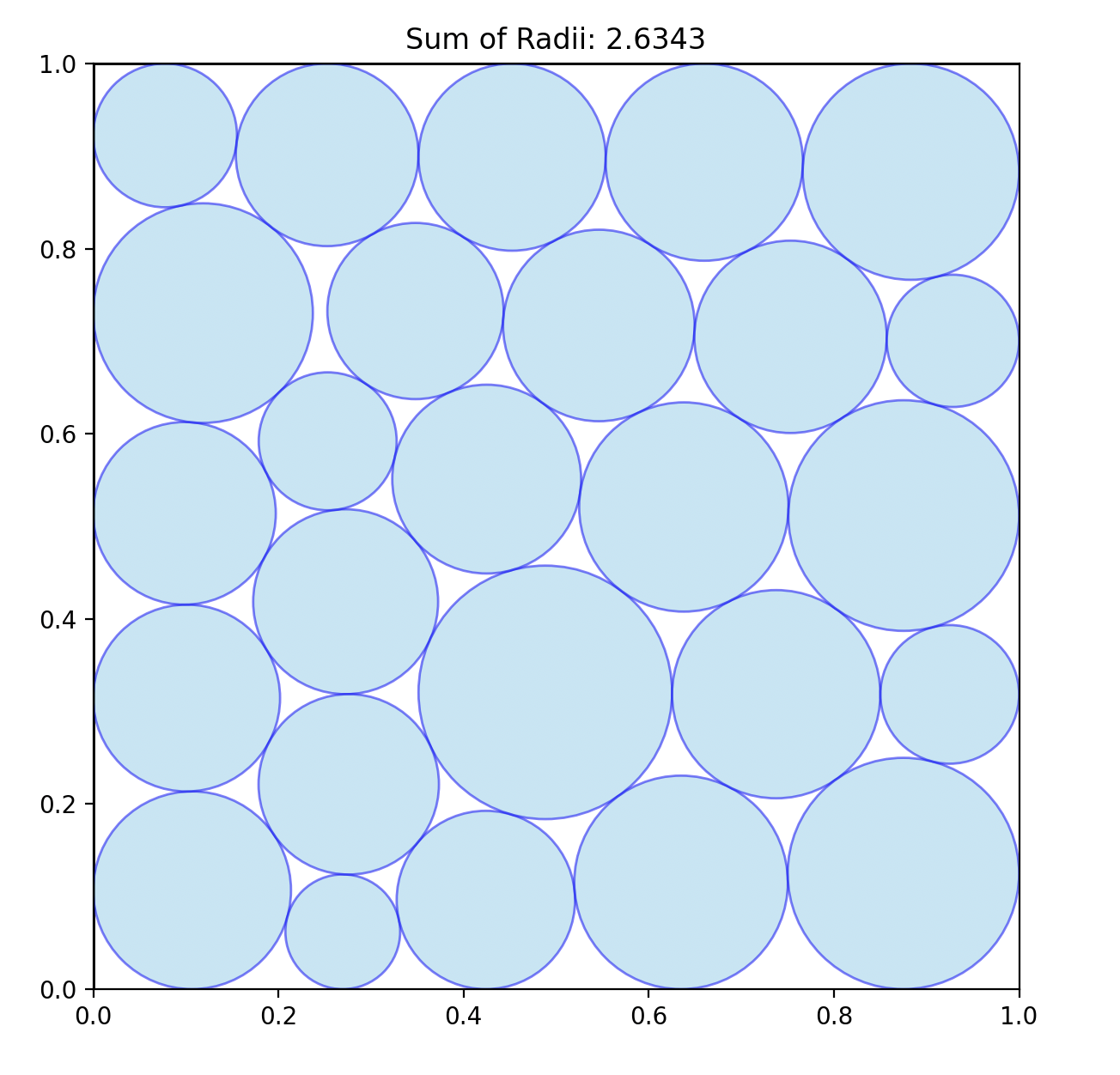}
        \caption*{OpenEvolve: Iteration 115}
    \end{subfigure}
    \caption{Visual comparison of circle packing solutions.}
    \label{fig:packing_comparison}
\end{figure}

\section*{3. New Discoveries}
In this section, we report four state-of-the-art discoveries, along with one highly competitive program, created by \name{}. Many of the evaluator programs and prompts used in these experiments are adapted from the TTT-Discover repository\cite{yuksekgonul2026learningdiscovertesttime}. We would like to thank them for making their evaluator scripts and prompts available.

See Appendix \ref{app:programs} for all programs \name{} created.

\subsection*{3.1 Erdős Minimum Overlap Problem}
The Erdős Minimum Overlap problem, originally posed by Paul Erdős in 1955, asks for a partition of the integers $\{1, \dots, 2n\}$ into two sets $A$ and $B$ such that the number of differences $a - b = k$ is minimized. In the continuous limit, this is equivalent to finding a function $f: [0, 2] \to [0, 1]$ with unit integral ($\int f = 1$) that minimizes the maximum value of its autocorrelation $(f * f)(t)$. The best human result on this is 0.380927.

We ran \name{}, passing in a highly basic initial program and an evaluator script adapted from OpenEvolve.\cite{openevolve} In 40 iterations, \name{} surpassed the TTT-Discover record (0.380876) to reach a new upper bound of \textbf{0.380874}. Notably, \name{}'s step function was much finer than the previous state-of-the-art's solution. While the previous state-of-the-art had 600 pieces, \name{}'s had 8192.

\begin{figure}[h!]
    
        \includegraphics[width=\linewidth]{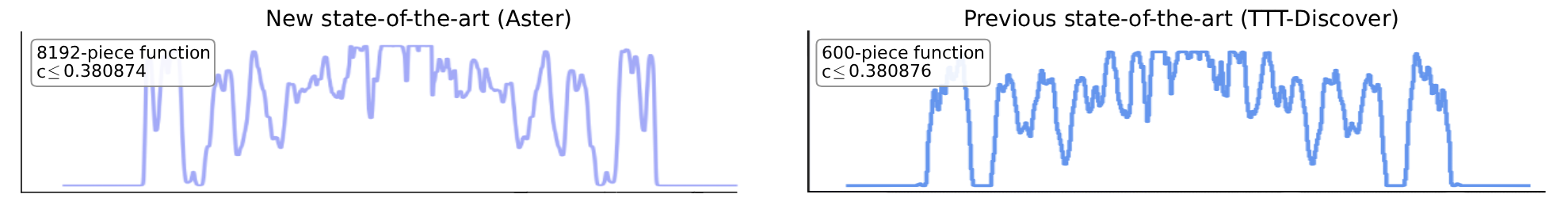}
    
    \caption{Comparison of Erdős Minimum Overlap constructions. Right: \name{}'s construction. Left: TTT-Discover.}
\end{figure}

\subsection*{3.2 Single-Cell Denoising}
Single-cell RNA sequencing (scRNA-seq) allows us to resolve biology at the level of individual cells, revealing cell types and states that bulk sequencing misses. However, this granularity comes at a cost: data is inherently sparse and noisy due to "dropout" events where expressed genes go undetected. Denoising algorithms are thus critical for recovering true gene expression profiles and maximizing the value of expensive sequencing experiments\cite{eraslan2019single}.

\name{} addressed this challenge on the OpenProblems benchmark\cite{luecken2025defining}, improving upon the TTT-Discover result (0.709) to reach \textbf{0.711} in 30 iterations. Our initial program was the best human script, and our evaluation criteria was strictly to minimize MSE while keeping Poisson below a certain threshold; this follows the molecular cross-validation framework established in \cite{batson2019molecular}.

\begin{table}[h!]
\centering
\small
\begin{tabular}{lccc}
\toprule
\textbf{Method} & \textbf{Mean Score} ($\uparrow$) & \textbf{MSE} ($\downarrow$) & \textbf{Poisson} ($\downarrow$) \\
\midrule
Best Human (MAGIC) & 0.641 & 0.190 & 0.050 \\
TTT-Discover & 0.709 & 0.154 & 0.048 \\
\midrule
\textbf{\textcolor{asterdark}{\name{}}} & \textbf{0.711} & \textbf{0.150} & \textbf{0.049} \\
\bottomrule
\end{tabular}
\caption{Single-cell denoising performance on the PBMC dataset. The Mean Score is the average of the normalized MSE and Poisson scores. \name{} achieves a higher overall mean score by significantly reducing MSE while maintaining competitive Poisson.}
\label{tab:denoising_results}
\end{table}

\begin{figure}[h!]
    \centering
    \includegraphics[width=0.8\linewidth]{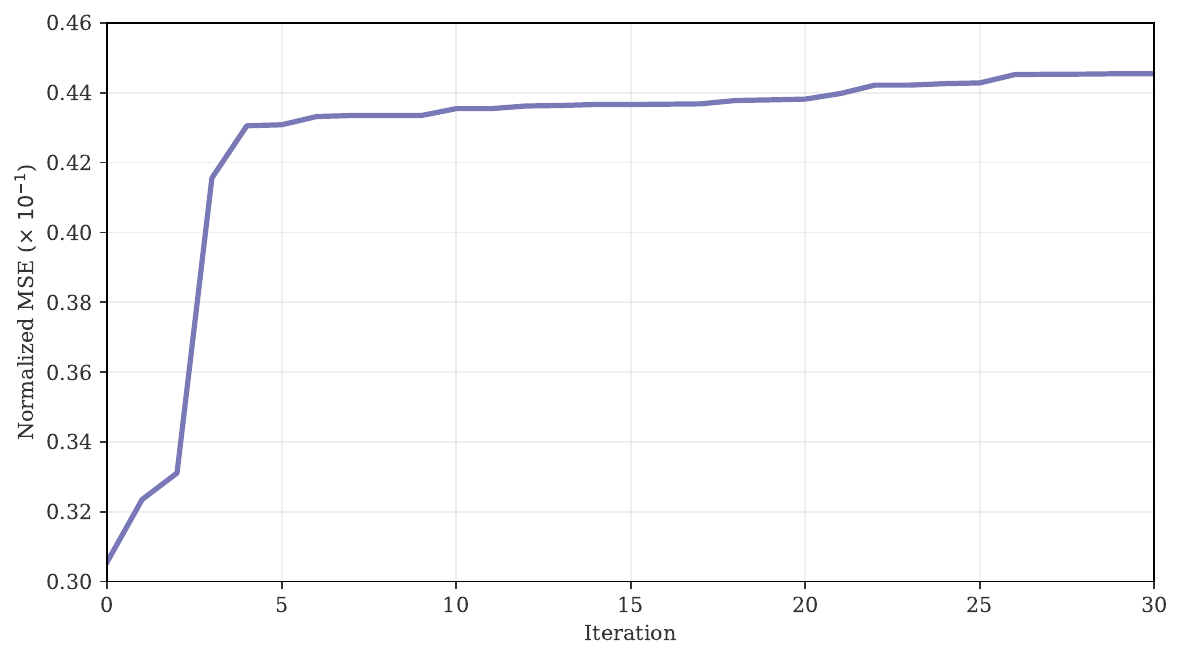}
    \caption{Trajectory of the best program found for Single-Cell Denoising.}
    \label{fig:denoising_curve}
\end{figure}

\subsection*{3.3: GPU Kernel Optimization}
GPU kernels are the computational foundation of modern AI; almost every training run at scale relies on highly optimized kernel code\cite{dao2022flashattentionfastmemoryefficientexact}. We targeted the forward pass of the Triangular Matrix Multiplication (TriMul) kernel, a core computational primitive in the AlphaFold architecture\cite{jumper2021highly} essential for protein structure prediction.

We set out to use \name{} to make the fastest version of the TriMul Kernel for the NVIDIA H100. We had previously done an optimization run for 94 iterations on a different GPU image than TTT-Discover, making the results difficult to compare. After discussing with the organizers of the TriMul competition, we found the correct image and restarted the optimization process. Our initial program was the best program found in the previous 94 iterations on this different image. Over 70 iterations, \name{} optimized the TriMul kernel's performance on an NVIDIA H100 GPU and was able to reduce runtime to \textbf{\SI{1114}{\micro\second}}. This outperforms the TTT-Discover benchmark of \SI{1161}{\micro\second}.

\begin{figure}[h!]
    \centering
    \includegraphics[width=0.8\linewidth]{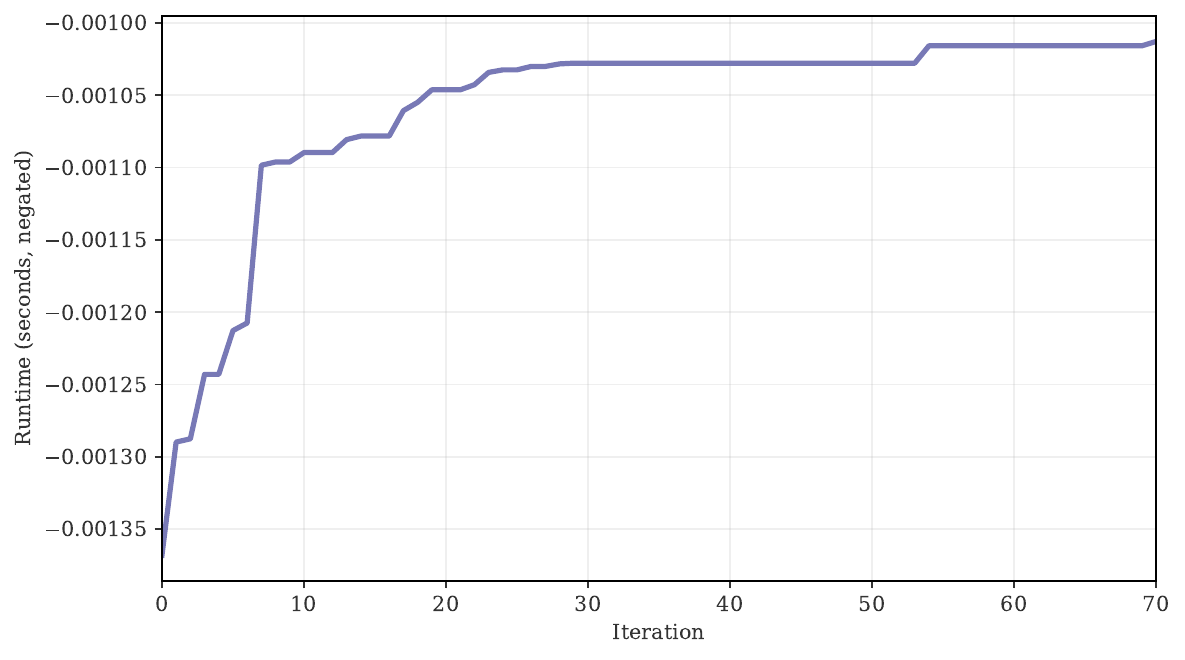}
    \caption{Trajectory of the best program found for GPU Kernel Optimization}
    \label{fig:gpu_curve}
\end{figure}
\subsection*{3.4: NanoGPT Speedrun Record}
A longstanding competition in machine learning is the NanoGPT Speedrun competition. This speedrun's purpose is to construct the fastest program that trains a language model with less than a 3.28 cross-entropy loss on the FineWeb validation dataset on a node of 8 NVIDIA H100 GPUs\cite{nanogpt_speedrun}. Advancements on this benchmark have led to several groundbreaking advancements in machine learning, most notably the Muon optimizer\cite{muon}, which was used to train Kimi K2\cite{kimi_k2}. When the competition started out, it took 45 minutes to train the model, and before Aster's submission, the record was 96.8 seconds.

In 8 iterations, \name{} shaved off 1.6 seconds to bring the record to 95.2 seconds.

\begin{table}[h]
\centering
\begin{tabular}{lr}
\toprule
\textbf{Method} & \textbf{Time (seconds)} \\
\midrule
Previous Best & 96.8 \\
\name{}'s Solution & 95.2 \\
\midrule
Speedup & 1.6\% \\
\bottomrule
\end{tabular}
\caption{Performance comparison on the NanoGPT Speedrun benchmark.}
\label{tab:nanogpt_results}
\end{table}

\begin{table}[h!]
\centering
\small
\begin{tabular}{lc}
\toprule
\textbf{AI System} & \textbf{Speedup} \\
\midrule
Locus\cite{locus} & 0.9\% \\
Hiverge\cite{hiverge} & 1.3\% \\
\midrule
\textbf{\textcolor{asterdark}{\name{}}} & \textbf{1.6\%} \\
\bottomrule
\end{tabular}
\caption{Comparison of AI-contributed speedups to the NanoGPT Speedrun record.}
\label{tab:nanogpt_ai_comparison}
\end{table}

\name{} is the third AI system to make a contribution to the NanoGPT Speedrun Record after Hiverge\cite{hiverge} and Locus\cite{locus}.

The solution that \name{} made here was a series of refinements to the Triton kernels in the program, optimizing the memory load-ins, and avoiding unnecessary recomputation.

\section*{3.5: ZAP-Bench}
All our previous tasks have had an evaluation time of at most a few minutes. To demonstrate the utility of Aster on tasks that take \textbf{hours to evaluate}, I chose the task of training a model on the Zebrafish Activity Prediction Benchmark (ZAP-Bench)\cite{zapbench}, a high-dimensional task requiring cellular-resolution forecasting of neural activity across an entire larval zebrafish brain. Specifically, I chose the short-context benchmark for predicting one step in the future, where the goal is to minimize the Mean Average Error (MAE).

The best human-created model for this task is a UNet architecture\cite{immer2025forecasting} which trained for 36 hours on 16 A100s. This model gets an MAE of 0.0182.

After 34 iterations (one 20-iteration run with a one-hour timeout, and a 14-iteration run with a 3-hour timeout), \name{} is able to create a model that also gets an MAE of 0.0182, matching the best human performance with 190x less compute. The evaluation script ran on an NVIDIA T4 GPU. This took \name{} approximately two and a half days of work. Without \name{}'s over 20x speedup, this evolution would likely have taken over a month.

Training a model on this task using an autonomous discovery system has already been done by Aygün et. al's tree search system \cite{aygün2025aihelpscientistswrite} and they were able to achieve an MAE of 0.0176 with similar runtime constraints. \name{} was still improving when we cut off its run, and it's highly likely that it would have been able to push the MAE down further with a longer runtime.

\begin{figure}[H]
    \centering
    \includegraphics[width=0.8\linewidth]{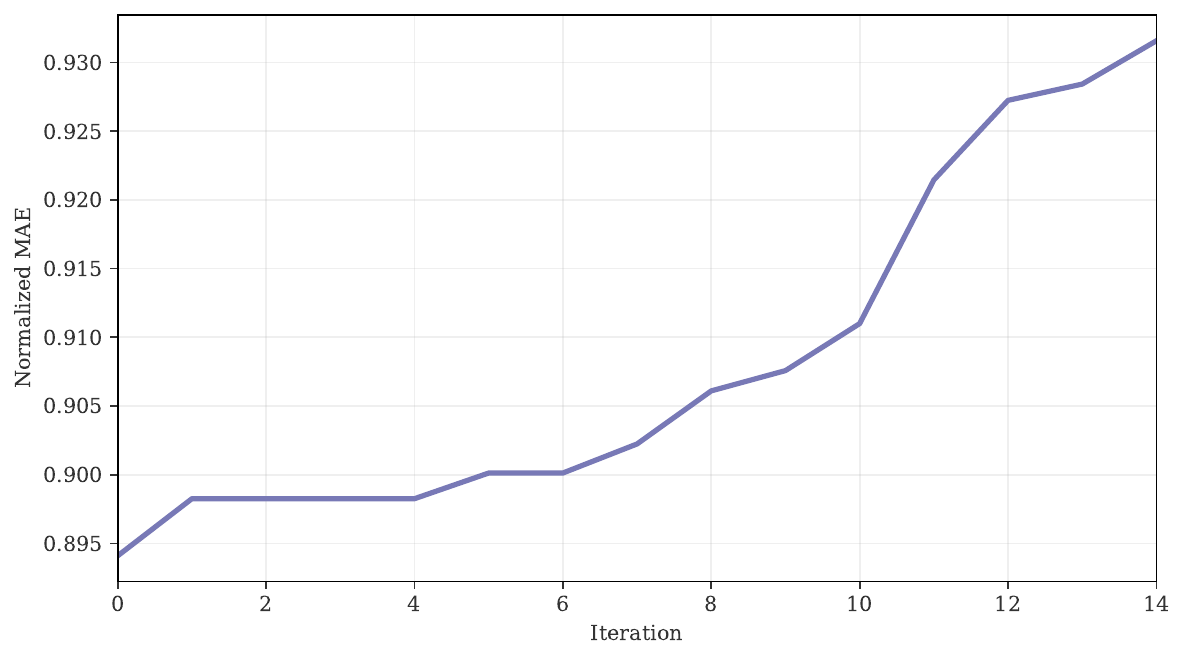}
    \caption{Trajectory of the best program found for ZAPBench forecasting. The normalized MAE is defined as 0.017/MAE.}
    \label{fig:zapbench_curve}
\end{figure}

\begin{figure}[H]
    \centering
    \includegraphics[width=0.6\textwidth]{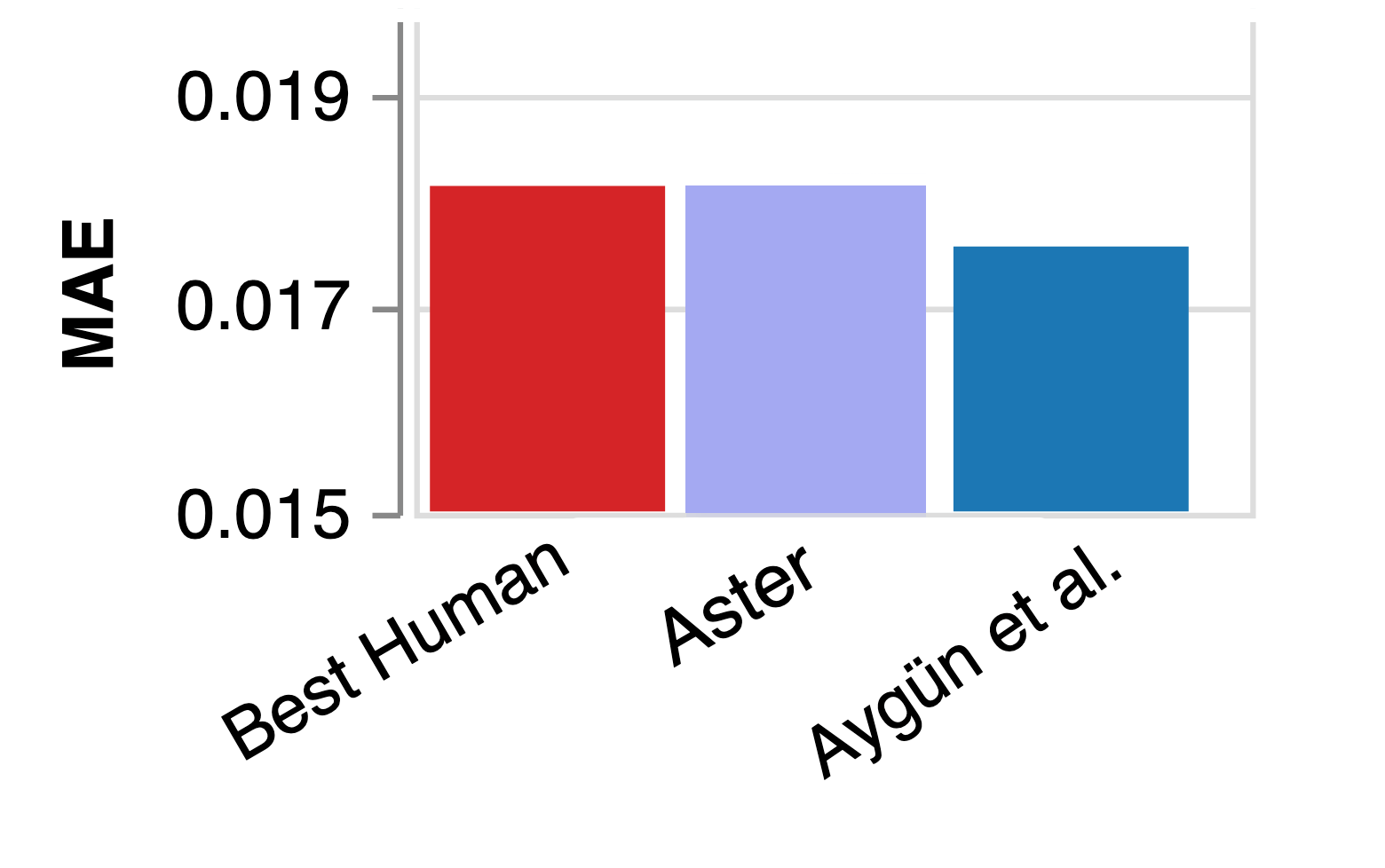}
    \caption{Comparison of ZAP-Bench short-context forecasting performance. \name{} matches the best human performance with significantly less compute.}
    \label{fig:zapbench_results}
\end{figure}

\bibliographystyle{plain}
\bibliography{references}

\appendix
\section{Programs}
\label{app:programs}

\tcbinputlisting{
    enhanced,
    breakable,
    listing only,
    colback=white,
    colframe=asterdark,
    colbacktitle=asterdark,
    fonttitle=\bfseries,
    coltitle=white,
    title={Erdős Minimum Overlap Program},
    boxrule=1pt,
    arc=1mm,
    listing file={results/erdos.py},
    listing options={
        style=artifactstyle,
        language=Python,
        basicstyle=\ttfamily\scriptsize,
        breaklines=true,
        keywordstyle=\color{blue},
        commentstyle=\color{gray},
        stringstyle=\color{teal},
    }
}

\tcbinputlisting{
    enhanced,
    breakable,
    listing only,
    colback=white,
    colframe=asterdark,
    colbacktitle=asterdark,
    fonttitle=\bfseries,
    coltitle=white,
    title={Single-Cell Denoising Program},
    boxrule=1pt,
    arc=1mm,
    listing file={results/denoising.py},
    listing options={
        style=artifactstyle,
        language=Python,
        basicstyle=\ttfamily\scriptsize,
        breaklines=true,
        keywordstyle=\color{blue},
        commentstyle=\color{gray},
        stringstyle=\color{teal},
    }
}

\tcbinputlisting{
    enhanced,
    breakable,
    listing only,
    colback=white,
    colframe=asterdark,
    colbacktitle=asterdark,
    fonttitle=\bfseries,
    coltitle=white,
    title={TriMul Kernel Optimization Program},
    boxrule=1pt,
    arc=1mm,
    listing file={results/trimul.py},
    listing options={
        style=artifactstyle,
        language=Python,
        basicstyle=\ttfamily\scriptsize,
        breaklines=true,
        keywordstyle=\color{blue},
        commentstyle=\color{gray},
        stringstyle=\color{teal},
    }
}

\tcbinputlisting{
    enhanced,
    breakable,
    listing only,
    colback=white,
    colframe=asterdark,
    colbacktitle=asterdark,
    fonttitle=\bfseries,
    coltitle=white,
    title={NanoGPT Speedrun Diff},
    boxrule=1pt,
    arc=1mm,
    listing file={results/nanogpt.diff},
    listing options={
        style=artifactstyle,
        language=Diff,
        basicstyle=\ttfamily\scriptsize,
        breaklines=true,
    }
}

\tcbinputlisting{
    enhanced,
    breakable,
    listing only,
    colback=white,
    colframe=asterdark,
    colbacktitle=asterdark,
    fonttitle=\bfseries,
    coltitle=white,
    title={ZAP-Bench Forecasting Program},
    boxrule=1pt,
    arc=1mm,
    listing file={results/zapbench.py},
    listing options={
        style=artifactstyle,
        language=Python,
        basicstyle=\ttfamily\scriptsize,
        breaklines=true,
        keywordstyle=\color{blue},
        commentstyle=\color{gray},
        stringstyle=\color{teal},
    }
}

\end{document}